\documentclass[runningheads]{llncs}
\usepackage{graphicx}
\usepackage{arydshln}
\usepackage{url}
\usepackage{amsmath}
\usepackage{booktabs}
\usepackage{multirow}
\usepackage{enumitem}
\usepackage{bm}
\begin{document}
\title{Coreference Resolution: Are the eliminated spans totally worthless?}
\author{Xin Tan* \and
Longyin Zhang \and
Guodong Zhou*}
\authorrunning{X. Tan et al.}
\institute{School of Computer Science and Technology, Soochow University, Suzhou, China \\
\email{\{xtan9,lyzhang9\}@stu.suda.edu.cn, gdzhou@suda.edu.cn}}
\maketitle              % typeset the header of the contribution
\begin{abstract}
Up to date, various neural-based methods have been proposed for joint mention span detection and coreference resolution. However, existing studies on coreference resolution mainly depend on mention representations, while the rest spans in the text are largely ignored and directly eliminated. In this paper, we aim at investigating whether those eliminated spans are totally worthless, or to what extent they can help improve the performance of coreference resolution. To achieve this goal, we propose to refine the representation of mentions with global spans including these eliminated ones leveraged. On this basis, we further introduce an additional loss term in this work to encourage the diversity between different entity clusters. Experimental results on the document-level CoNLL-2012 Shared Task English dataset show that the eliminated spans are indeed useful and our proposed approaches show promising results in coreference resolution.

\end{abstract}
\section{Introduction}
\renewcommand{\thefootnote}{}
\footnotetext{* Corresponding author}
As an important role in text understanding, coreference resolution is the task of identifying and clustering mention spans in a text into several clusters where each cluster refers to the same real world entity.
With the increasing population of neural networks, varied neural-based approaches have been proposed for coreference resolution~\cite{wiseman-etal-2016-learning,clark2016deep,clark-manning-2016-improving,lee-etal-2017-end,zhang-etal-2018-neural-coreference,lee-etal-2018-higher,kantor-globerson-2019-coreference,fang2019incorporating,joshi-etal-2019-bert,fei-etal-2019-end,wu-etal-2020-corefqa} so far.
Among these studies, Lee et al.\cite{lee-etal-2017-end} first propose an end-to-end neural model apart from syntactic parsers. After that, the coreference resolution task is much liberated from complicated hand-engineered methods, and more and more studies~\cite{zhang-etal-2018-neural-coreference,lee-etal-2018-higher,kantor-globerson-2019-coreference,fang2019incorporating,joshi-etal-2019-bert} have been proposed to refine the coreference resolution model of Lee et al.~\cite{lee-etal-2017-end}.
In general, neural-based studies usually perform coreference resolution in two stages: (i) a mention detector to select mention spans from all candidate spans in a document and (ii) a coreference resolver to cluster mention spans into corresponding entity clusters. In this way, the manually annotated cluster labels are well employed for both mention detection and clustering.

Although previous studies have achieved certain success in coreference resolution in recent years, these studies perform mention clustering heavily rely on the mention representations selected at the first stage for mention clustering, while the rest spans in each piece of text are largely filtered and directly eliminated. Considering that, we naturally raise the question: Are those eliminated spans worthless for coreference resolution?
On the one hand, the mention spans selected by the mention scoring function are usually isolated with each other in format. Besides, eliminating these ``worthless'' spans only depending on the mention scoring function may aggravate the notorious error propagation problem in the pipeline workflow of coreference resolution. On the other, previous studies improve the performance of coreference resolution either from feature designing or model architecture perspectives, while the progress of improving on data utilization remains hysteretic. As far as we know, these eliminated spans have not been explored so far. Under this condition, we aim at increasing the data utilization rate and investigating to what extent these eliminated spans can help improve the performance of coreference resolution.

To achieve the above goal, based on the two-stage neural model of Lee et al.~\cite{lee-etal-2017-end}, we propose a mention representation refining strategy to well leverage the spans that highly related to the mention for representation enhancing.
Following this way, the contribution of our approach is two-fold: (i) using the global spans (both mention spans and the eliminated spans) with high utilization rate to provide auxiliary information for mention representation enhancing; (ii) equipping the isolated mention representations with context-aware correlations through the trade-off among global spans in the document.
In addition, to make full use of the annotated training instances, we utilize an additional loss term to learn from both positive and negative samples to encourage the diversity between different entity clusters. Notably, we also explore the effects of two different contextualized word embeddings (i.e., ELMo~\cite{peters-etal-2018-deep} and BERT~\cite{devlin2018bert:}) in this paper for comparison. Experimental results on the document-level CoNLL-2012 English dataset show that our way of reusing these eliminated spans is quite useful for the coreference resolution task, and our approach shows promising results when compared with previous state-of-the-art methods.

\section{Background}

\subsubsection{Task Definition.} Following previous studies~\cite{lee-etal-2017-end,zhang-etal-2018-neural-coreference,lee-etal-2018-higher,kantor-globerson-2019-coreference,fang2019incorporating,joshi-etal-2019-bert,fei-etal-2019-end}, we cast the task of coreference resolution as an antecedent selection problem, where each span is assigned with an antecedent in the document. Specifically, given a span $i$, the possible antecedents are $Y_i = \{\epsilon, 1, 2,..., i-1\}$ (\textit{i.e.}, a dummy antecedent $\epsilon$ and all preceding spans). The non-dummy antecedent refers to the coreference link between span $i$ and its antecedent $y_i$ ($y_i\neq\epsilon$). The dummy antecedent denotes the coreference link between span $i$ and $\epsilon$, which represents two possible scenarios: (i) span $i$ is not an entity mention or (ii) span $i$ is an entity mention but not coreferent with any previous span.

\subsubsection{Baseline.}
In this work, the baseline model~\cite{lee-etal-2017-end} we use for coreference resolution learns a distribution $P(y_i)$ over the antecedents of each span $i$:
\begin{equation}
    P(y_i)=\frac{e^{S(i,y_i)}}{\sum_{y' \in Y(i)}e^{S(i,y')}}
\end{equation}
where $S(i, j)$ represents a pairwise score for the coreference link between span $i$ and span $j$.
And the pairwise coreference score $S(i, j)$ is calculated based on the mention scores of $i$ and $j$ (i.e., $S^m_i$ and $S^m_j$ which denote whether the spans $i$ and $j$ are mentions or not) and the joint compatibility score of $i$ and $j$ (i.e., $S^a_{i,j}$ which denotes whether mention $j$ is an antecedent of mention $i$ or not). And the final pairwise coreference score is written as:
\begin{equation}
  S(i, j)=S^m_i+S^m_j+S^a_{i,j}
\end{equation}

Given the vector representation $g_i$ for each possible span $i$, the mention score of span $i$ and the antecedent score between spans $i$ and $j$ can be calculated as:
\begin{gather}
  S^m_i = W_m \cdot \text{FFNN}_m(g_i) \\
  S^a_{i, j} = W_a \cdot \text{FFNN}_a([g_i, g_j, g_i \circ g_j, \phi(i, j)])
\end{gather}
where $g_i$ is obtained via bidirectional LSTM models that learn context-dependent boundary and head representations, $W_m$ and $W_a$ denote two learnable parameter matrixes, $\circ$ denotes the element-wise multiplication, $\text{FFNN}(\cdot)$ denotes a feed-forward neural network, and the antecedent score $S^a_{i, j}$ is calculated through explicit element-wise similarity of each span, $g_i \circ g_j$, and a feature vector $\phi(i, j)$ that encodes speaker and genre information from the metadata and the distance between spans $i$ and $j$.

\section{Coreference Resolution with Enhanced Mention Representation}

Motivated by previous studies~\cite{lee-etal-2017-end,zhang-etal-2018-neural-coreference,fang2019incorporating}, we inherit the architecture that combines mention detection and coreference scoring for coreference resolution. Particularly, as stated before, we propose to reuse the spans that are eliminated at the mention detection stage to enhance the representation of mention spans for better coreference resolution performance.

\subsection{Mention Detection} At the mention detection stage, we take the text spans within a certain length limitation as potential mention spans.
Following previous studies, we take word-, character- and context-level information for span representation~\cite{lee-etal-2017-end}. Moreover, we also incorporate the syntactic structural information (e.g. the path, siblings, degrees, and category of the current node) for representation enhancing~\cite{fang2019incorporating}:
\begin{equation}
  \textbf{s}_i = [\bm{h}_{bi}, \bm{h}_{ei}, \hat{\bm{x}}_i, \bm{f}_i]
\end{equation}
where $\bm{h}_i$ is the contextual representation of input $x_i =[w_i, c_i]$ ($w_i$ means the word embedding vector and $c_i$ means the character embedding vector). $\hat{\bm{x}}_i$ is the weighted sum of the word representations that contained in the span $i$, where the attention weights are learned through the Head-finding attention mechanism, and one can refer to~\cite{lee-etal-2017-end} for the detailed process of calculating $\hat{\bm{x}}_i$. And $\bm{f}_i$ denotes the syntactic structural feature vector.

After that, we use a feedforward scoring function for mention determination as Lee et al.~\cite{lee-etal-2017-end} did:
\begin{equation}
  S^m_i = W_m \cdot \text{FFNN}_m(\textbf{s}_i)
\end{equation}
Then, the spans that assigned with high attention scores are selected as the resulting mention spans, noted by $\textbf{m}_i$.

\subsection{Coreference Resolving with Global Spans Perceived} 
\begin{figure*}[t]
  \begin{small}
  \begin{center}
    \includegraphics[scale=0.9]{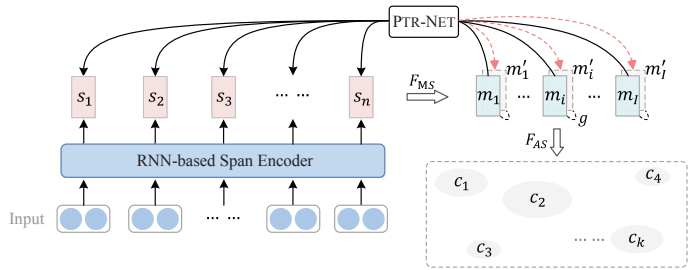}
    \caption{\label{fg_example1}Mention representation refining for coreference resolution. Here, $F_{MS}$ and $F_{AS}$ refer to the mention scoring function and antecedent scoring function respectively, and $c_k$ denotes an independent entity cluster.}
  \end{center}
  \end{small}
\end{figure*}

After obtaining the mention representation, the following coreference scoring stage aims to determine the antecedent for each mention. In order to reduce the problem of error propagation in the pipeline workflow of coreference resolution, we propose to further enhance the mention representation with context-aware correlations. To achieve this, we propose to extract information from both mentions and all the eliminated spans to help refine the obtained mention representations, as illustrated in Figure~\ref{fg_example1}.

Concretely, we use a pointer net (PTR-NET)~\cite{vinyals2015pointer} in this work to extract the context-aware information from the spans that are strongly related to the current mention. More formally, we extract context information and project it into a parallel mention vector space as:
\begin{gather}
  u_{i,t} = V^T {\rm tanh}(W_1\textbf{s}_t + W_2\textbf{m}_i) \\
  \alpha_i = {\rm softmax}(\textbf{u}_i) \\
  \textbf{m}'_i = \sum \alpha_{i,t} \cdot \textbf{s}_t, t \in \{1,...,n\}
\end{gather}
where $\textbf{s}_t$ denotes the representation of the $t$-th span, $\textbf{m}_i$ refers to the representation of the $i$-th mention, $V^T$, $W_1$ and $W_2$ are learnable parameter matrixes, $\alpha_{i,t}$ denotes the relevancy of each span, and $\textbf{m}'_i$ denotes the context-aware representation for the mention $i$ in the parallel mention vector space.
In this manner, (i) the spans that strongly related to the current mention are well leveraged to exploit context information and (ii) the correlations between eliminated spans and mentions are well learned through the trade-off among spans in the text during the attention calculation process.

After achieving the parallel mention representations $\textbf{m}_i$ and $\textbf{m}'_i$, we add a gated model between them for information communication, corresponding to $g$ in Figure~\ref{fg_example1}. And the enhanced mention representation is formulated as:

\begin{gather}
  f_i = \delta(W_f[\textbf{m}_i, \textbf{m}'_i]) \\
  \textbf{m}^*_i = f_i \circ \textbf{m}_i + (1-f_i) \circ \textbf{m}'_i
\end{gather}
where $f_i$ is a learnable vector for information filtering, $\circ$ denotes the element-wise multiplication, and $\textbf{m}^*_i$ refers to the integrated mention representation.

Then, the pairwise antecedent score is calculated based on the integrated mention pairs. Following previous work~\cite{lee-etal-2017-end}, the coreference score is generated by summing up the mention score and the pairwise antecedent score as:

\begin{gather}
  S^a_{i,j}=W_a\cdot\text{FFNN}_a([\textbf{m}^*_i,\textbf{m}^*_j,\textbf{m}^*_i\circ\textbf{m}^*_j,\!\phi(i,\!j)]) \\
  S(i, j)=S^m_i+S^m_j+S^a_{i,j}
\end{gather}
where $S^a_{i,j}$ denotes the antecedent score between spans $i$ and $j$, and $S(i, j)$ denotes the final coreference score.

\section{Model Training}

Previous studies usually cluster candidate mentions only depending on gold cluster labels. Differently, based on the loss objective of Lee et al.~\cite{lee-etal-2017-end}, we introduce an additional loss term to maximize the distance between different entity clusters to encourage the diversity. Concretely, we make the loss term learn from both positive and negative samples, and the loss objective can be formulated as:

\begin{gather}
    \mathcal{L} = - {\rm log} \prod_{i=1}^{N}\sum_{\hat y \in Y(i) \cap \text{GOLD}(i)} P(\hat y) + {\rm log} \prod_{i=1}^{N}\sum_{\hat y \notin Y(i) \cap \text{GOLD}(i)} P(\hat y)
\end{gather}
\begin{equation}
  P(\hat y)=\frac{e^{S(i,\hat y)}}{\sum_{y' \in Y(i)}e^{S(i,y')}}
\end{equation}
where $S(i,\hat y)$ denotes the coreference score of the coreference link between span $i$ and its antecedent $\hat y$, and $\text{GOLD}(i)$ denotes the set of entity mentions in the cluster that contains span $i$. If span $i$ does not belong to any clusters, or if all gold antecedents have been pruned, $\text{GOLD}(i)$ equals $\{\varepsilon\}$.
Through the proposed approach, the training instances are well utilized to cluster entity mentions and at the same time increase the diversity between different mention clusters.

\section{Experimentation}

\subsection{Experimental Settings}

\subsubsection{Datasets.} We carry out several experiments on the English coreference resolution data from the CoNLL-2012 Shared Task~\cite{pradhan-etal-2012-conll}.
The dataset contains 2802 documents for training, 343 documents for validation, and 348 documents for testing with 7 different genres (i.e., newswire, magazine articles, broadcast news, broadcast conversations, web data, conversational speech data, and the New Testament). Experimental data: \url{https://cemantix.org/conll/2012/data.html}.

\subsubsection{Model Settings.} We used three kinds of word representations for experimentation, i.e., (i) the fixed 300-dimensional GloVe~\cite{pennington2014glove:} vectors and the 50-dimensional Turian~\cite{turian2010word} vectors, (ii) the 8-dimensional character embeddings learned from CNNs, where the window sizes of the convolution layers were 3, 4, and 5 characters respectively, and each layer consists of 50 filters, and (iii) two kinds of 1024-dimensional contextualized word representations provided by ELMo~\cite{peters-etal-2018-deep} and BERT~\cite{devlin2018bert:}.
The model hyper-parameters were directly borrowed from Lee et al.~\cite{lee-etal-2017-end} for fair comparison.

\subsubsection{Metrics.} We report the Precision, Recall, and F$_1$ scores based on three popular coreference resolution metrics, i.e., $MUC$~\cite{vilain-etal-1995-model}, $B^{3}$~\cite{bagga1998algorithms}, and $CEAF_{\phi4}$~\cite{luo-2005-coreference}. And we report the averaged F$_1$-score as the final CoNLL score.

\subsection{Experimental Results}

In this paper, we select the model of Lee et al.~\cite{lee-etal-2017-end} as our baseline system.
We borrow the system implemented by Kong and Fu~\cite{fang2019incorporating} for experimentation in two reasons:
(i) Kong and Fu~\cite{fang2019incorporating} incorporate varied syntactic structural features (e.g., the path, siblings, degrees, and category of the current node) into the model of Lee et al.~\cite{lee-etal-2017-end} to better capture hierarchical information for span representation. (ii) Their implemented system can well reduce computational complexity whose training speed is 6 times faster than that of Lee et al.~\cite{lee-etal-2017-end}.
Moreover, we enhance their system by applying the ELMo embedding to it for better comparison.
We report the results of the original and enhanced baseline systems of Kong and Fu~\cite{fang2019incorporating} for performance comparison. Moreover, following previous work, we also present the performances of recent systems for reference, and the overall results are shown in Table~\ref{tab:Table 1}.

\begin{table*}[t]
\small
\centering
\begin{tabular}{lcccccccccc}
\toprule
   & \multicolumn{3}{c}{$\text{MUC}$} &  \multicolumn{3}{c}{$\text{B}^3$} &  \multicolumn{3}{c}{$\text{CEAF}_{\phi4}$} \\
  \cmidrule(r){2-4} \cmidrule(r){5-7} \cmidrule(r){8-10}
  &Prec. & Rec. & F$_1$   & Prec. & Rec. & F$_1$   & Prec. & Rec. & F$_1$  & Avg. F$_1$ \\
\midrule
Wu et al.~\cite{wu-etal-2020-corefqa} $\ddagger$ & 88.6  & 87.4  & 88.0  & 82.4  & 82.0 & 82.2  & 79.9  & 78.3  & 79.1 & 83.1 \\
\cdashline{1-11}[0.8pt/2pt]
Wiseman et al.~\cite{wiseman-etal-2016-learning}  & 77.5 &  69.8 &  73.4 &  66.8 &  57.0 &  61.5 &  62.1 &  53.9 &  57.7 & 64.2 \\
Clark and Manning~\cite{clark-manning-2016-improving}  & 79.9 &  69.3 &  74.2 &  71.0 &  56.5 &  63.0 &  63.8 &  54.3 & 58.7 & 65.3 \\
Clark and Manning~\cite{clark2016deep}  & 79.2 &  70.4 &  74.6 &  69.9 &  58.0 &  63.4 &  63.5 &  55.5 &  59.2 & 65.7 \\
Lee et al.~\cite{lee-etal-2017-end}  & 78.4  & 73.4	 & 75.8  & 68.6  & 61.8  & 65.0  & 62.7  & 59.0  & 60.8  & 67.2 \\
Lee et al.~\cite{lee-etal-2018-higher} $\dagger$  & 81.4  & 79.5  & 80.4  & 72.2  & 69.5  & 70.8  & 68.2  & 67.1  & 67.6  & 73.0 \\
Zhang et al.~\cite{zhang-etal-2018-neural-coreference}  & 79.4  & 73.8  & 76.5  & 69.0  & 62.3  & 65.5  & 64.9  & 58.3  & 61.4  & 67.8 \\
\hline
Baseline~\cite{fang2019incorporating}  & 80.5  & 73.9  & 77.1  & 71.2  & 61.5  & 66.0  & 64.3  & 61.1  & 62.7  & 68.6 \\
Baseline~\cite{fang2019incorporating} $\dagger$  & 80.8	 & 79.3	 & 80.0  & 71.2  & 68.7	 & 70.0 	& 66.4	& 66.3	& 66.4	& 72.1 \\
Ours $\dagger$  & \textbf{81.8}	& 80.0 	& \textbf{80.9}	& \textbf{72.6}	& 70.0 	& \textbf{71.3}	& \textbf{68.1}	& \textbf{67.9}	& \textbf{68.0} & \textbf{73.4} \\
~ -\textit{additional\_loss}  & 80.9	 & \textbf{80.6} 	& 80.8	& 70.6	& \textbf{70.6} 	& 70.6	& 67.2	& \textbf{67.9}	& 67.5 	& 73.0 \\
\bottomrule
\end{tabular}
\caption{Performance comparison on coreference resolution. Sign ``$\dagger$'' means the ELMo representation is used and ``$\ddagger$'' means the powerful Bert model is employed. Compared with the enhanced baseline system, our performance improvements on F$_1$ are statistically significant with $p<0.05$.}
\label{tab:Table 1}
\end{table*}

From the results we can find that (i) Comparing our approach (line 10) with the baseline systems (lines 8 and 9), our method performs better on all the three metrics, which suggests the great effectiveness of our proposed method in utilizing all spans including the eliminated ones for coreference resolution. And the contextualized ELMo also helps improve the performance to a certain extent.
(ii) Comparing our system with previous state-of-the-art methods, ours outperforms most of the systems except the Bert-based model of Wu et al.~\cite{wu-etal-2020-corefqa}.
(iii) Comparing our system under different model settings (the last two lines) we find that the additional loss term we use can improve the performance of coreference resolution to some extent, especially on the $B^{3}$ indicator.
The overall results above indicate that our proposed methods are useful and can well increase the utilization rate of the coreference resolution data.

\subsection{Analysis on Context-Aware Word Representations}

In this subsection, we present our insight on exploring a better pre-trained word representation. More theoretically, we aim to figure out a question: Which kind of context-aware embeddings is better for coreference resolution?
With this in mind, we employ two kinds of popular contextualized word representations (i.e., ELMo~\cite{peters-etal-2018-deep} and BERT~\cite{devlin2018bert:}) for analysis. Briefly review:
\begin{itemize}[leftmargin=*]
  \item \textbf{ELMo.} Peters et al.~\cite{peters-etal-2018-deep} hold the view that word representation should contain rich syntactic and semantic information and be able to model polysemous words. On this basis, they provide the contextualized ELMo for word representation by training bidirectional LSTMs on a large-scale corpus.
  \item \textbf{BERT.} Devlin et al.~\cite{devlin2018bert:} present two new pre-training objectives, i.e., the ``masked language model (LM)'' for word-level representation and the ``next sentence prediction'' for sentence-level representation. In essence, Bert achieves a good feature representation of words by running the self-supervised learning method on the basis of massive training data.
\end{itemize}

\begin{figure*}[t]
  \begin{small}
    \begin{center}
      \includegraphics[scale=0.32]{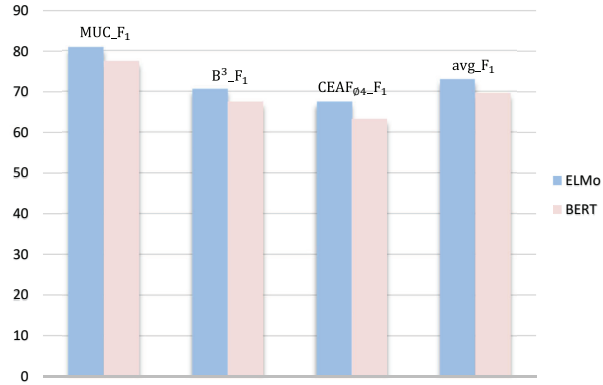}
      \caption{\label{fig:embed} Performance comparison between ELMo and Bert on coreference resolution.}
    \end{center}
  \end{small}
\end{figure*}
Comparing the two kinds of contextualized word embeddings above: (i) ELMo is better for feature-based methods, which transfers specific downstream NLP tasks to the process of pre-training to produce the word representations, so as to obtain a dynamic word representation that changes constantly according to the context; (ii) Bert prefers fine-tuning methods, which fine-tunes network parameters to produce better word representation for specific downstream NLP tasks.
Previous Bert-based work~\cite{wu-etal-2020-corefqa} has proven the great effectiveness of fine-tuning the Bert model in coreference resolution.
Different from them, in this work, we analyze the effects of static features extracted by pre-trained LMs.
To achieve this, we employ ELMo and Bert models to generate static context-aware embeddings and then apply them to the coreference resolution task.
Performance comparison between the two kinds of word vectors are detailed in Figure~\ref{fig:embed}.

Usually, the context features generated by ELMo are known to be shallower than that generated by Bert. But the results show that the Bert-based coreference resolution system performs relatively worse than the ELMo-based one.
In other words, the shallow-level features defeat deep-level ones in our coreference resolution experiments. One possible reason for this result is that the ELMo-based shallow context features are more easily understood and utilized by the resolution system while the Bert-based deep context features could be too much for the model to use directly. Nevertheless, fine-tuning the Bert model can transform the informative and complicated context information into a task-specific one for better understanding, which explains the recent success in fine-tuning Bert for better coreference resolution~\cite{wu-etal-2020-corefqa}.

\subsection{Case Study}

\begin{figure*}
  \begin{small}
  \begin{center}
    \includegraphics[scale=0.95]{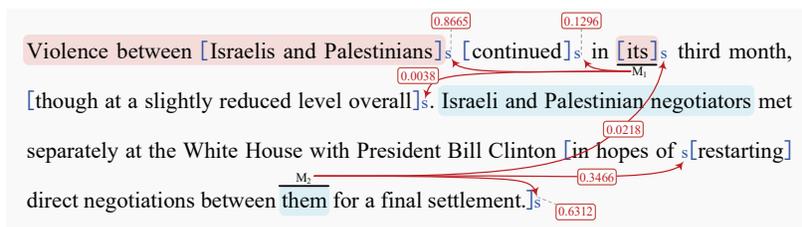}
    \caption{\label{fg_example2}Example of the mention-related span selection process, where text fragments of the same color (``Violence ...'' and ``its'' are colored in orange; ``Israeli ...'' and ``them'' are colored in blue) belong to the same cluster. And the solid lines denote the scores of the most possible three spans assigned for each mention. }
  \end{center}
  \end{small}
\end{figure*}

To qualitatively analyze the mention representation refining process, we provide a visualization of the span pointing process, as shown in Figure~\ref{fg_example2}. From the example, there exit two entity clusters with different background colors. Here, we present the top three spans that are particularly relevant to each mention (i.e., $M_1$ and $M_2$) for reference. Obviously, both mentions, $M_1$ and $M_2$, are pronouns and they pay much attention to those spans that semantically related to them (e.g., Violence between Israelis and Palestinians). However, most of these spans with high scores are eliminated during the mention detection stage in previous studies, which much hinders the interaction between mentions and mention-related spans. Fortunately, in this work, the pointer mechanism~\cite{vinyals2015pointer} is well utilized to detect these mention-related spans, and the selected spans are reused for mention representation refining. It is worth noting that mention representation refining is effective especially for pronoun mentions with ambiguous meanings, e.g., the two mentions \textit{its} and \textit{them} in the example. And the proposed method can give these mentions more accurate context-aware representation for better coreference resolution.

\section{Related Work}

Before the population of neural networks, traditional machine learning methods have a long history in the coreference resolution task. In general, three mainstream methods have been proposed: 1) Mention-pair models~\cite{ng-cardie-2002-improving,bengtson-roth-2008-understanding} to determine if a pair of mentions are coreferent by training binary classifiers. 2) Mention-ranking models~\cite{denis2007ranking,wiseman-etal-2015-learning,clark2016deep} to score all previous candidate mentions for current mention and select the most possible one as its antecedent. 3) Entity-mention models~\cite{clark-manning-2015-entity,clark-manning-2016-improving,wiseman-etal-2016-learning} to determine whether the current mention is coreferent with a preceding, partially-formed mention cluster. Although these methods have achieved significant performance gains, they still suffer from one major drawback: their complicated hand-engineered rules are difficult to adapt to new languages.

With the rapid spread of neural network in recent years, varied researchers turned to neural-based methods.
Lee et al.~\cite{lee-etal-2017-end} proposed the first end-to-end neural-based system that liberates coreference resolution from the complicated hand-engineered methods.
Zhang et al.~\cite{zhang-etal-2018-neural-coreference} proposed to improve the performance of coreference resolution by using a biaffine attention model to score antecedents and jointly optimize the two sub-tasks.
Lee et al.~\cite{lee-etal-2018-higher} proposed an approximation of higher-order inference using the span-ranking architecture from Lee et al.~\cite{lee-etal-2017-end} in an iterative manner.
Kong and Fu~\cite{fang2019incorporating} proposed to improve the resolution performance by incorporating various kinds of structural information into their model.
Kantor and Globerson~\cite{kantor-globerson-2019-coreference} proposed to capture properties of entity clusters and use them during the resolution process.
Most recently, Fei et al.~\cite{fei-etal-2019-end} presented the first end-to-end reinforcement learning based coreference resolution model.
Joshi et al.~\cite{joshi-etal-2019-bert} and Wu et al.~\cite{wu-etal-2020-corefqa} proposed to improve the performance of coreference resolution with the help of the state-of-the-art Bert.

\section{Conclusion}

In this paper, we aim at increasing the data utilization rate and exploring the value of those spans eliminated at the mention detection stage. On this basis, we proposed a mention representation refining strategy where spans that highly related to the mentions are well leveraged through a pointer network for representation enhancing. Moreover, we introduced an additional loss term to encourage the diversity between different entity clusters. Notably, we also performed experiments on different contextualized word embeddings to explore the effectiveness of them on coreference resolution.
Experimental results on the document-level CoNLL-2012 Shared Task English dataset indicate that these eliminated spans are indeed useful and our proposed approach can achieve much better results when compared with the baseline systems and competitive results when compared with most previous studies in coreference resolution.

\section*{Acknowledgements}

This work was supported by the National Natural Science Foundation of China (NSFC) via Grant Nos. 62076175, 61876118, and the Postgraduate Research \& Practice Innovation Program of Jiangsu Province under Grant No. KYCX20\_2669. Also, we would like to thank
the anonymous reviewers for their insightful comments.
\bibliographystyle{splncs04}
\bibliography{nlpcc2021}

\end{document}